%% file: main.tex
\documentclass[runningheads]{llncs}

\usepackage{eccv}

\usepackage{eccvabbrv}

\usepackage{graphicx}
\usepackage{booktabs}

\usepackage[accsupp]{axessibility}  

\usepackage{hyperref}

\usepackage{orcidlink}
\usepackage{pifont}

\usepackage{listings}
\usepackage{xcolor} 

\begin{document}

\title{WebRPG: Automatic Web Rendering Parameters \\ Generation for Visual Presentation} 
\titlerunning{Automatic Web Rendering Parameters Generation for Visual Presentation}

\author{Zirui Shao\inst{1}\thanks{\ Equal contribution.}\orcidlink{0000-0002-4210-070X} \and
Feiyu Gao\inst{2}$^\star$\orcidlink{0009-0009-3206-5347} \and
Hangdi Xing\inst{1}\orcidlink{0000-0002-1770-005X} \and
Zepeng Zhu\inst{1}\orcidlink{0009-0000-1510-6455} \and
Zhi Yu\inst{1}\thanks{\ Corresponding author.}\orcidlink{0009-0001-8608-5628} \and \\
Jiajun Bu\inst{1}\orcidlink{0000-0002-1097-2044} \and
Qi Zheng\inst{2}\orcidlink{0009-0001-3822-2616} \and
Cong Yao\inst{2}\orcidlink{0000-0001-6564-4796}}


\authorrunning{Z. Shao et al.}

\institute{Zhejiang Provincial Key Laboratory of Service Robot, Zhejiang University \\
\email{\{shaozirui, xinghd, zapeng, yuzhirenzhe, bjj\}@zju.edu.cn} \and
Alibaba Group \\
\email{feiyu.gfy@alibaba-inc.com, yongqi.zq@taobao.com, yaocong2010@gmail.com}}

\maketitle

\setcounter{footnote}{0}

\begin{abstract}
    In the era of content creation revolution propelled by advancements in generative models, the field of web design remains unexplored despite its critical role in modern digital communication. The web design process is complex and often time-consuming, especially for those with limited expertise. In this paper, we introduce Web Rendering Parameters Generation (WebRPG), a new task that aims at automating the generation for visual presentation of web pages based on their HTML code. WebRPG would contribute to a faster web development workflow. Since there is no existing benchmark available, we develop a new dataset for WebRPG through an automated pipeline. Moreover, we present baseline models, utilizing VAE to manage numerous elements and rendering parameters, along with custom HTML embedding for capturing essential semantic and hierarchical information from HTML. Extensive experiments, including customized quantitative evaluations for this specific task, are conducted to evaluate the quality of the generated results. The dataset and code can be accessed at GitHub\footnote{\url{https://github.com/AlibabaResearch/AdvancedLiterateMachinery/tree/main/DocumentUnderstanding/WebRPG}}.
    
\end{abstract}

\input{2_Introduction}
\input{Related_Work}
\input{3_Dataset}

\input{4_Methodology} 
\input{5_Experiment}
\input{6_Conclusion}

\section*{Acknowledgements}
This work is supported by the National Natural Science Foundation of China (Grant No. 62372408) and the National Key R\&D Program of China (No. 2021YFB2701100).


%
%
\bibliographystyle{splncs04}
\bibliography{main}

\clearpage
\input{X_suppl}

\end{document}

%% file: 2_Introduction.tex
\section{Introduction}

Recently, we are witnessing a revolution in content creation, driven by rapid advancements in generative models across domains such as image \cite{stable_diffusion,ramesh2022hierarchical,ddpm,nichol2021improved, vqvae1}, text \cite{GPT3,Touvron2023Llama2O,li2022diffusion}, and audio \cite{kong2020diffwave,chen2021wavegrad,wavegrad2}. 
Numerous studies aim to leverage these advancements to enhance efficiency in graphic design, including advertisement~\cite{li2020attribute,NDN} and magazine \cite{magazine,NDN,zheng2019content} design. Nevertheless, the automation of web design, an essential part of graphic design \cite{wang2022influence}, lacks exploration. Web design plays a significant role in the visual communication of web pages \cite{aesthetics_web}, impacting not only user satisfaction \cite{cyr2010colour} but also user behavior \cite{flavian2009web}. 
Yet, it is a complex, time-consuming task, especially challenging for those developers with limited design expertise, leading to substandard visual presentations~\cite{williams2015non}. Automating web design can simplify this process, enabling developers to create visually appealing web pages, and bridging the gap between technical development and aesthetic excellence.

Web pages are formed by HTML\footnote{\url{https://html.spec.whatwg.org/}} and CSS\footnote{\url{https://www.w3.org/Style/CSS/specs.en.html}} code, where HTML defines the content and structure, and CSS controls the visual presentation. With the advent of large language models (LLMs) \cite{Touvron2023Llama2O,GPT3,Ouyang2022TrainingLM}, automating HTML code generation has become feasible. However, efforts in automatic visual presentation design, the core aspect of web design, currently center on specific subtasks such as layout generation \cite{kikuchi2021modeling,web_layout2,web_layout3}, font recommendation \cite{zhao2018modeling,font2}, and colorization \cite{kikuchi2023generative,qiu2022intelligent,gu2016data}, rather than designing a holistic web visual presentation from scratch.

\begin{figure}[t]
\centering
\includegraphics[width=0.99\columnwidth]{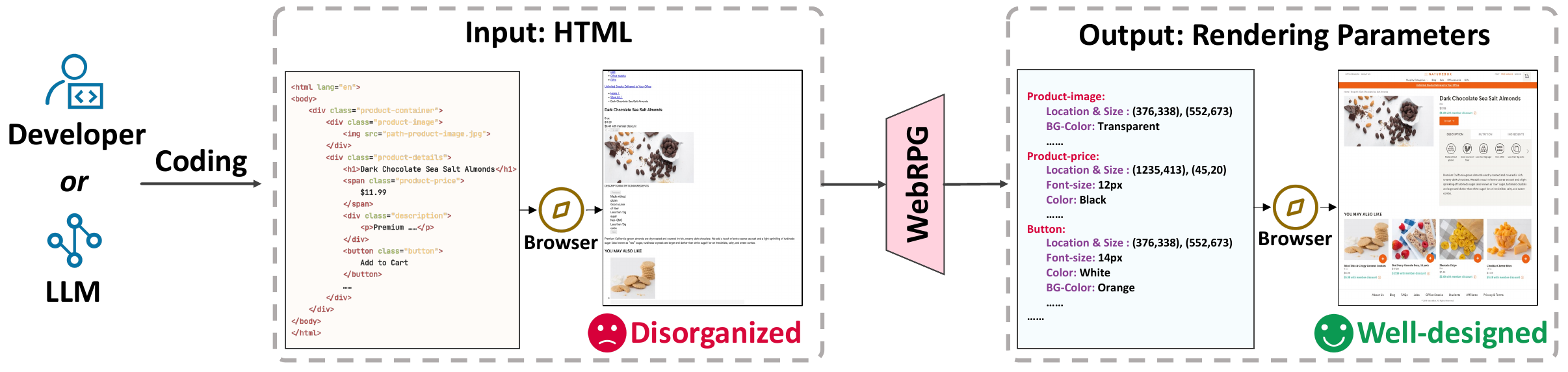} 
\caption{Overview of the WebRPG task. The input consists of plain HTML code and the output comprises rendering parameters for each element. With browser rendering, plain HTML produces a disorganized visual presentation, while incorporating the generated rendering parameters significantly enhances the visual presentation.}
\label{fig:toyexample}
\end{figure}

Intuitively, leveraging generative models to learn design knowledge from existing web pages is a practical strategy for automated web visual design. However, the complexity of CSS coding practices poses challenges for its automatic generation \cite{kaluarachchi2023systematic}.  To address this, we propose standardizing CSS using Rendering Parameters (RPs), which are defined by CSS properties that control the visual appearance of each web element \cite{css}. Consequently, we introduce a novel task called \textbf{Web} \textbf{R}endering \textbf{P}arameters \textbf{G}eneration (\textbf{WebRPG} for short), which requires the automatic generation of rendering parameters for each web element based on the HTML code, as depicted in \cref{fig:toyexample}. 
With the help of a WebRPG system, HTML is the only prerequisite for obtaining an effective web visual presentation, which has the potential to achieve a faster web development workflow. With the integration of LLMs, a WebRPG system can even enable the realization of a fully automated web development workflow. Moreover, it can facilitate new applications, such as efficient exploration of various design options and dynamic personalization of web page styles.

Since there is no existing benchmark available for WebRPG, we develop automatic data processing steps to transform raw web pages into formalized WebRPG samples and construct a new dataset utilizing the Klarna dataset~\cite{hotti2021klarna}. From a theoretical perspective, the WebRPG task presents two primary challenges:  1) Web pages comprise hundreds of elements, each with numerous RPs. 2) The visual presentation of web elements should be associated with the semantic and hierarchical information provided by HTML code. To address the challenges, variational autoencoder (VAE) \cite{Kingma2013AutoEncodingVB} is employed to handle the large volume of rendering parameters for web elements, and specially designed HTML embedding is introduced to encode semantic and hierarchical information from HTML code. Using these modules, two WebRPG baselines are established, which are based on autoregressive and diffusion models, respectively. To verify the effectiveness of WebRPG baselines, metrics are designed to evaluate the overall appearance, layout, and style of the generated results. Both quantitative and qualitative experiments are conducted to assess the baselines.

Our main contributions are as follows:

\begin{itemize}
    \item 
    We introduce a novel task WebRPG for automatic web design from HTML code and create a new dataset.
    \item We explore the WebRPG task by establishing two baselines and propose solutions for its challenges.
    \item We design metrics to quantitatively evaluate the quality of generated results, and conduct qualitative experiments to analyze the strengths and weaknesses of the baselines.
\end{itemize}

%% file: Related_Work.tex
\section{Related Work}
Generative models achieve notable success in image \cite{stable_diffusion,ramesh2022hierarchical,maskgit,vqvae1,vqvae2}, text \cite{GPT3,Touvron2023Llama2O,li2022diffusion}, and audio \cite{kong2020diffwave,chen2021wavegrad,wavegrad2}. Image synthesis can create web visual presentations by generating screenshots but struggles with producing coherent text \cite{ramesh2022hierarchical}. Moreover, image synthesis is limited to static images and cannot offer interactive, manipulable web pages.

Numerous efforts utilize generative models for graphic design, including advertising \cite{li2020attribute,NDN}, magazines \cite{magazine,zheng2019content,hui2023unifying}, UI \cite{inoue2023layoutdm,jyothi2019layoutvae,play,liu2018learning}, and posters\cite{cgl,9525300}. Yet, the designs restrict the element count to no more than 25. These methods primarily employ a one-dimensional sequence to represent designs, with each element defined by five tokens: four describe the bounding box, and one indicates the category (e.g., text, headline) \cite{NDN}. However, the reliance on a simplistic flat input for the WebRPG task, which involves managing hundreds of elements and various RPs, leads to a substantial memory consumption increment, and performance degradation \cite{Dong2023ASO}. Moreover, the one-dimensional sequence neglects crucial hierarchical information in web pages.

Research focused on web pages has continuously emerged. In terms of understanding, efforts in web question answering \cite{websrc,Zhao2022TIETI}, web information extraction  \cite{swde,Xie2021WebKEKE}, and web pre-trained language models \cite{Li2021MarkupLMPO,dom-lm,shao2023gem} have made notable progress in comprehending the essential semantic content and hierarchical structure of web pages. For instance, MarkupLM \cite{Li2021MarkupLMPO} stands out with its unique architecture and pre-training tasks, effectively encoding HTML content, which offers insights for our research. Moreover, there are works aimed at web page design, such as optimizing the overall or specific block coloring of web pages \cite{kikuchi2023generative,qiu2022intelligent,gu2016data}, determining layouts based on given components like navigation bars \cite{kikuchi2021modeling,web_layout2,web_layout3}, and recommending fonts for particular elements \cite{zhao2018modeling,font2}. However, these studies focus only on specific subtasks of the web page design workflow, leaving the comprehensive design of web pages from scratch as an unexplored area.

%% file: 3_Dataset.tex
\section{Preliminary}

\subsection{Task Definition}
\label{sec:task_fotmula}
Web design is centered on visual presentation, i.e., the manipulation of CSS code. The complexity of CSS coding practices, including a wide range of selector options, makes the automatic generation of CSS code challenging \cite{kaluarachchi2023systematic}. To facilitate the model for learning web design, we standardize CSS by converting it into rendering parameters (RPs), which can be transformed back into CSS,  with additional details in Sec. A.1. Consequently, the WebRPG task is defined as follows: given the HTML code, generate rendering parameters for each web element. Specifically, given a web page $\mathcal{X}$, whose HTML code is $\mathcal{H}$, it consists of a set of elements $\mathcal{X}=\{X_1,X_2,\ldots,X_S\}$, where $S$ is the number of elements in $\mathcal{X}$. The visual appearance of element $X_i$ is controlled by a set of RPs denoted as $P_i = \{p_i^k \mid k \in \mathcal{W}\}$, where $\mathcal{W}$ indicates the indices for all RPs, and the complete set of RPs for $\mathcal{X}$ is $\mathcal{P} = \{P_1, P_2,\ldots, P_S\}$. Therefore, the primary objective of the WebRPG task is to create a function $f$ that generates RPs based on HTML code, that is, $f: (\mathcal{H}) \mapsto \mathcal{\hat P}$, where $\mathcal{\hat P}$ represents the estimate of $\mathcal{P}$.

\subsection{Web Rendering Parameter Definition}
\label{sec:rp_definition}

The term ``\textbf{Rendering Parameters (RPs)}'' is employed to collectively describe the parameters controlling the visual appearance of each web element on the browser, as defined by CSS properties. Layout and visual style are crucial in the design of web pages~\cite{Segmentation_Gestalt,shao2023gem}, leading us to summarize 13 common CSS properties, divided into 3 categories as follows.

\begin{itemize}
    \item \textbf{Layout properties} include \textit{left}, \textit{top}, \textit{width}, and \textit{height}.
    \item \textbf{Text properties} include \textit{font-style}, \textit{font-weight}, \textit{font-size}, \textit{line-height}, \textit{text-align}, \textit{text-decoration}, and \textit{text-transform}.
    \item \textbf{Color properties} include \textit{color} and \textit{background-color}.
\end{itemize}

Various formats are available for web developers to define CSS properties. To standardize, we adopt the values computed by the browser \cite{MDN2023} as the reference. Specifically, the values related to position and size are uniformly measured in integer pixels, and the values related to color correspond to 46 widely used colors. The vocabulary for all rendering parameters is available in Sec. A.1.

\section{Dataset Construction}
\subsection{Data Pre-processing}
\label{sec:data_pre_process}
Raw web pages cannot provide straightforward supervision for RPs. Thus, several pre-processings are conducted. \textit{Headless chrome}\footnote{\url{https://developer.chrome.com/blog/headless-chrome/}} is used to render web pages and \textit{selenium}\footnote{\url{https://www.selenium.dev/}} is employed to store HTML with only visible elements and record each element’s selected CSS properties. Note that elements in this paper mean nodes in the DOM\footnote{\url{https://www.w3.org/DOM/DOMTR}} tree. The elements are stored following the DOM tree's pre-order traversal. Since many web pages retain thousands of elements, we treat elements with a certain number of children as sub-pages with the semantic and hierarchical integrity preserved. The sub-pages are further cleaned while keeping the visual appearance, including removing uncommon HTML tags and intricate components like carousel images, as well as placing sub-pages at the top-left corner of the browser. Additionally, we only consider static components. Our models disregard the image on web pages, preserving only \(\textless\)img\(\textgreater\) tags. 
To guarantee data quality, a specific Visual Complexity (VC) metric is introduced to assist in filtering samples. The metric integrates three dimensions: color, size, and alignment, inspired by previous works \cite{alemerien2014guievaluator,fu2007measuring}. The definition of the VC metric is provided in Sec. A.2.

\begin{figure}[t]
\centering
\includegraphics[width=0.95\columnwidth]{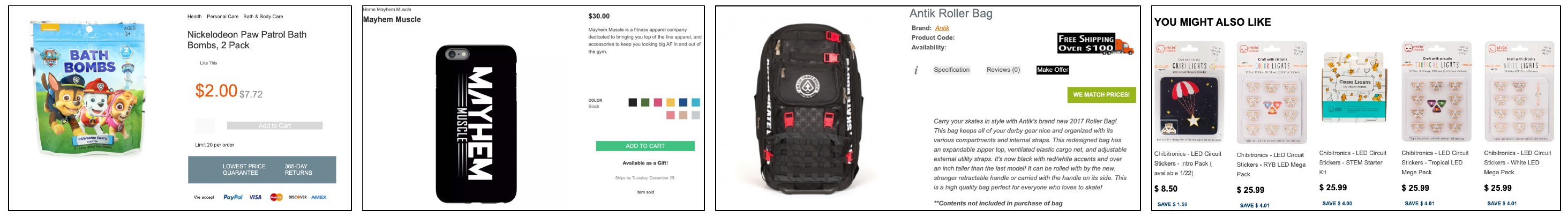} 
\caption{Selected sub-page screenshots from our dataset. Notably, regions displayed are cropped due to space limitations.}
\label{fig:data_sample}
\end{figure}

\subsection{Dataset Details}
\label{sec:our_dataset}
To accommodate the requirement for offline rendering, the Klarna dataset \cite{hotti2021klarna} is utilized to build our WebRPG dataset. The Klarna dataset, initially used for web information extraction, comprises 20K English product pages from 3K e-commerce sites, ensuring domain-specific diversity. The dataset stores all pages in MHTML\footnote{\url{https://en.wikipedia.org/wiki/MHTML}} format, enabling offline rendering of the original pages in browsers with high fidelity. 

The pre-processing in \cref{sec:data_pre_process} is applied to the web pages with the browser canvas size setting to 1920*1920 pixels, generating sub-pages containing between 32 and 128 child elements. The token length for each sample (sub-page) does not surpass 512. The size of RP vocabulary is 1993. The samples with a VC below 0.1 are filtered out. After preprocessing, our dataset includes 88,418 samples, split into training and testing sets at an 8:2 ratio. Our dataset exceeds the size of established graphic design datasets such as CLAY \cite{li2022learning} (50K samples) and RICO \cite{liu2018learning} (43K samples), ensuring it can meet our objectives. Screenshots of some samples are shown in \cref{fig:data_sample}. More details are provided in Sec. A.3.

%% file: 4_Methodology.tex
\begin{figure}[t]
\centering
\includegraphics[width=0.95\textwidth]{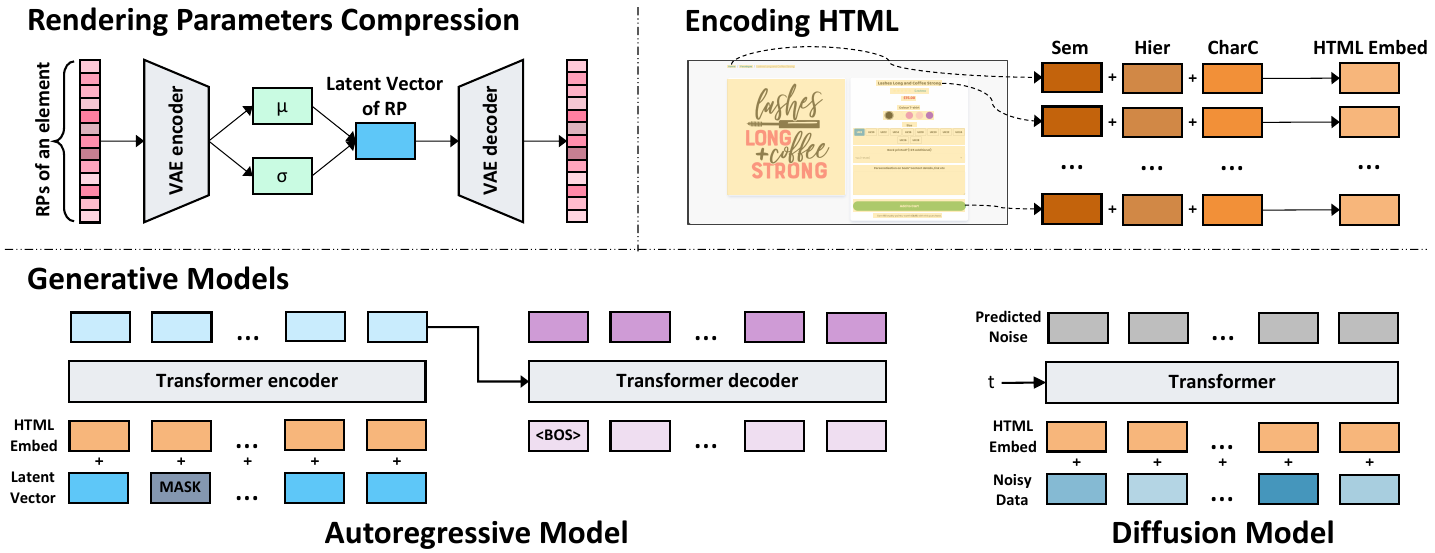} 
\caption{Key components of WebRPG models. In the upper left, VAE compresses the RPs of each element into latent vectors shown in \textcolor[RGB]{95,199,248}{\textbf{blue}}. In the top right, "Semantic" (Sem), "Hierarchical" (Hier), and "Character Count" (CharC) embeddings combine into the HTML embedding in \textcolor[RGB]{247,182,124}{\textbf{orange}}. Below, two generative models are illustrated.}
\label{fig:model}
\end{figure}

\section{Methodology}
\label{sec:method}
\subsection{Overview}
As indicated in \cref{sec:task_fotmula}, the WebRPG task is formulated as a function that generates rendering parameters (RPs) for each web element based on the HTML code. Inspired by classical generation methods \cite{stable_diffusion,vqvae2,vqvae1}, we employ a latent generation approach. In the approach, VAE is leveraged to compress all RPs of an element into latent space representation (\cref{sec:vae}), and a generative model (\cref{sec:core_model}) generates the latent vector based on the given HTML embeddings (\cref{sec:html_embed}), which is then decoded back into RPs by the decoder of VAE. The key components of our method are shown in \cref{fig:model}.

\subsection{Rendering Parameters Compression}
\label{sec:vae}
Assume a web page consists of $S$ elements, with the appearance of each element $X_i$ determined by $\mathcal{W}$ rendering parameters $P_i = \left\{p_i^k \mid k \in \mathcal{W}\right\}$. The WebRPG model necessitates the processing of $S \times \mathcal{W}$ values for both input and output. Expanding all $p_i^k$ of $X_i$ into a one-dimensional sequence, as per graphic design methods \cite{inoue2023layoutdm,NDN}, leads to excessively long input and output lengths. To mitigate this challenge, we utilize VAE to compress the rendering parameters into a latent space. This ensures that the input length for the generative model correlates solely with \( S \).

More precisely, given the RPs of an element $P_i \in \mathbb{R}^{\mathcal{W}*V}$, where $V$ is the size of RPs vocabulary (\cref{sec:rp_definition}), and the corresponding latent vector is $Z_i \in \mathbb{R}^{d}$. We denote the generative distribution as $p_{\theta}(P_i \mid Z_i)$ and the posterior as $q_{\phi}(Z_i \mid P_i)$, respectively. The learning objective of VAE is expressed as:
\begin{equation}
  L_{VAE}= \frac{1}{S} \cdot \sum_{i=1}^S   ( - \mathbb{E}_{q_{\phi}(Z_i \mid P_i)}\left[\log p_{\theta}(P_i \mid Z_i)\right] +\lambda_{KL} \mathrm{KL}\left(q_{\phi}(Z_i \mid P_i) \|  p(Z_i) \right) ),
\label{eq:vae_leaning}
\end{equation}
where $\theta$ and $\phi$ are the encoder and decoder parameters, $\mathbb{E}$ indicates the expectation,  
 $\mathrm{KL}$ is the Kullback-Leibler divergence, and $\lambda_{KL}$ is the hyperparameter to balance the two terms. The encoder and decoder of VAE both consist of a multilayer perceptron with five layers. To ensure that the latent space encompasses as many element appearances (i.e., combinations of RPs) as possible, the VAE is pre-trained using synthetic data.

\subsection{Encoding HTML}
\label{sec:html_embed}
The visual presentation of a web page should be in harmony with the content and structure dictated by its HTML code. To this end, we design an \textbf{HTML embedding} that captures the essential information in the HTML code, establishing the input feature for the generative model (\cref{sec:core_model}). HTML code essentially encompasses hierarchical information among elements and the textual content of each element~\cite{dom-lm}. The character count of each element is also crucial, as the size of an element generally exhibits a positive correlation with the length of characters. Therefore, our HTML embedding integrates three facets of information: semantics, hierarchy, and character count. Precisely, for an element \(X_i\), its HTML embedding  \( H_i \in \mathbb{R}^d\) is defined as:
\begin{equation}
   H_i = \Lambda^{\text{Sem}}(H_{i}^{\text{Sem}}) + \Lambda^{\text{Hier}}(H_{i}^{\text{Hier}}) + \Lambda^{\text{CharC}}(H_{i}^{\text{CharC}}),
\end{equation}
where \(H_{i}^{\text{Sem}}\), \(H_{i}^{\text{Hier}}\) and \(H_{i}^{\text{CharC}}\) denote the semantic, hierarchical and character count embedding respectively, and $\Lambda^{\circ}()$ is the linear projection layer.

\textbf{Semantic embedding: }The MarkupLM$_{large}$ model~\cite{Li2021MarkupLMPO}, a language model explicitly pre-trained for web understanding, is employed as the semantic extractor. Specifically, given an element $X_i$ with HTML code tokens $X_i = \{x_i^j \mid j \in \mathcal{L} \}$, where $\mathcal{L}$ denotes the token length, we calculate the semantic embedding of \(X_i\) as \(H_{i}^{\text{Sem}} = \text{Pool}(\text{MarkupLM}(x_i^1, x_i^2, \ldots, x_i^\mathcal{L}))\), where \(\text{Pool}(\cdot)\) denotes an average pooling operation.

\textbf{Hierarchical embedding:} The XPath embedding layer~\cite{Li2021MarkupLMPO} is employed to model the hierarchical information of elements, taking their XPath expressions as input. XPath\footnote{\url{https://www.w3.org/TR/xpath-31/}} is a query language for selecting elements from a web page, which is based on the DOM tree and can be used to easily locate an element. Specifically, for an element \( X_i \) with its corresponding XPath expression \( xp_i \), we compute the hierarchical embedding directly as \( H_{i}^{\text{Hier}} = \text{XPathEmb}(xp_i) \).

\textbf{Character count embedding:} We establish a mapping mechanism that translates the raw count of characters into dense vector space. For an element $X_i$ with the content of $k$ characters, the character count embedding is calculated as \(H_{i}^{\text{CharC}} = \text{EmbCharC}(k)\).

\subsection{Generative Models}
\label{sec:core_model}
Two generative models are implemented: autoregressive and diffusion model.

\textbf{Autoregressive Model (AR):}
\label{sec:mle_model}
To enhance the model stability during training, a masked latent vector  \(\mathcal{Z}_{mask}\) of real RPs is introduced inspired by BART \cite{bart} and MaskGIT \cite{maskgit}. \(\mathcal{Z}_{mask}\) is constructed in two steps. Firstly, the real RPs are encoded into the latent vectors with the VAE encoder, i.e., \(\mathcal{Z} = \theta(\mathcal{P})\). Then a special \(MASK\) vector and a binary mask \(M = \{m_i \mid i \in S\}\) are utilized to partially substitute the real latent vectors with the \(MASK\) as \(Z_{mask,i} = m_i \cdot MASK + (1 - m_i) \cdot \theta(P_i)\).

Here \(M\) is generated using a mask scheduling function \(\gamma(r) \in (0,1]\) following MaskGIT \cite{maskgit}, and the $MASK$ vector is a learnable parameter with the same shape as \(Z_i\). Additionally, it is important to highlight that during inference, all \( Z_i \) are masked, i.e., \( M = \{m_i = 1 | 1 \leq i \leq S\} \).

As depicted in \cref{fig:model}, the model inputs the sum of \(\mathcal{Z}_{mask}\) and \(\mathcal{H}\) to generate \(\hat{\mathcal{Z}}\), which is then decoded by the VAE decoder as \(\mathcal{\hat{P}} = \phi(\hat{\mathcal{Z}})\).  The VAE and generative models are trained jointly, thus the training loss is as follows:
\begin{equation}
    L = \log p_{\psi}(\mathcal{P} | \mathcal{H},\mathcal{Z}_{mask}) +  L_{VAE},
\end{equation}
where \(\psi\) is the parameters of the generative model.

\textbf{Diffusion Model:} Diffusion models \cite{ddpm,nichol2021improved,zhu2022discrete} have recently emerged as a new class of generative models with high performance. These models are characterized by forward and reverse Markov processes of length $T$. In our rendering parameters compression (VAE) model, rendering parameters $\mathcal{P}$ are encoded into a latent space, i.e., $\mathcal{Z} = \theta(\mathcal{P})$. These latent vectors $\mathcal{Z}$, which align more closely with a Gaussian distribution, improve compatibility with the noise distribution in diffusion models. Following successful models \cite{dhariwal2021diffusion,stable_diffusion,saharia2022image}, our diffusion model can be interpreted as an equally weighted sequence of denoising autoencoders $\mathcal{E}(\mathcal{Z}_t,t,\mathcal{H}); t = 1 \ldots T$, which are trained to predict the noise $\boldsymbol{\epsilon} \sim \mathcal{N}(\mathbf{0}, \mathbf{I})$ in $\mathcal{Z}_t$. The $\mathcal{Z}_t$ is obtained from a forward process starting from $\mathcal{Z}_0$ (where $\mathcal{Z}_0=\mathcal{Z}$), defined as $\mathcal{Z}_t = \sqrt{\alpha_t} \mathcal{Z}_{t-1} + \sqrt{1-\alpha_t}\boldsymbol{\epsilon}$, with $\alpha_t$ being a predefined set of coefficients. As illustrated in \cref{fig:model}, $\mathcal{Z}_t$ and $\mathcal{H}$ are added and input into the model. Our diffusion model employs the standard variational lower bound objective as its training loss, and we jointly optimize the VAE, leading to the overall loss function:
\begin{equation}
L= \mathbb{E}_{\mathcal{Z},\epsilon\sim\mathcal{N}(0,1),t}\Big[\|\boldsymbol{\epsilon}-\boldsymbol{\epsilon}_\psi(\mathcal{Z}_t,t,\mathcal{H})\|_2^2\Big] +  L_{VAE}.
\end{equation}
During inference, the predicted $\mathcal{\hat{Z}}$ is progressively obtained through a reverse process, expressed as $\mathcal{Z}_{t-1} = \frac{1}{\sqrt{\alpha_t}} \left( \mathcal{Z}_t - \frac{1-\alpha_t}{\sqrt{1-\alpha_t}} \boldsymbol{\epsilon}_\psi(\mathcal{Z}_t, t, \mathcal{H}) \right)$. Subsequently, $\mathcal{\hat{Z}}$ is decoded to $\mathcal{\hat{P}}$ via a single pass through the VAE decoder $\phi$. Additionally, $\mathcal{Z}_T$ is random Gaussian noise.

%% file: 5_Experiment.tex
\section{Experiment}
\subsection{Evaluation Metrics}
\label{sec:metric}
Three metrics are utilized to assess the quality of the generated rendering parameters. Fréchet Inception Distance (FID), Element Intersection over Union (Ele. IoU), and newly introduced Style Consistency Score (SC Score) enable the evaluation of the overall appearance, layout, and style of generated web pages respectively. As indicated in \cref{sec:rp_definition}, ``layout'' refers to layout properties, while ``style'' encompasses text properties and color properties.

\subsubsection{Fréchet Inception Distance}
\label{sec:fid}

FID \cite{heusel2017gans}, a metric initially proposed in the domain of image generation, measures the similarity of generated data to real ones in feature space. Inspired by Lee et al. \cite{NDN}, a binary classifier is trained to distinguish between real and noise-added RPs. This classifier is employed to generate representative features of RPs for calculating FID. We also introduce layout-specific and style-specific FID models. Further details are in Sec. A.4.

\subsubsection{Elements Intersection over Union}
Ele. IoU is a metric for evaluating the similarity between generated layouts and real ones, based on adaptation to the Maximum IoU \cite{kikuchi2021constrained}. As the elements of real and generated web pages correspond one-to-one, IoU is computed between the corresponding pairs. Denote the real layouts as \( B = \{b_i\}_{i=1}^{N} \) and the generated ones as \( \hat{B} = \{\hat{b}_i\}_{i=1}^{N} \), with \( N \) being the element count, and \( b_i \) and \( \hat{b}_i \) as corresponding elements. The Ele. IoU can be calculated as follows:
\begin{equation}
    \text{EleIoU}(B, \hat{B}) = \frac{1}{N} \sum_{i=1}^{N} IoU(b_i, \hat{b}_i).
\end{equation}

\subsubsection{Style Consistency Score}
\label{sec:sc_score}
The ``Principle of Similarity'' of Gestalt theory suggests that people tend to perceive elements with similar style as a whole \cite{gestalt,gestaltprinciples}, highlighting the importance of style consistency among elements. Hence, the SC Score assesses whether elements with the same style on a real web page retain that consistency on the generated page, beyond merely visual similarity. An example explanation is provided in Sec. C. Elements are deemed to have the same style only if all their style properties are identical \cite{shao2023gem}. Specifically, for a web page \(W = \{e_i \mid i \in N\}\) with \(N\) being the number of elements, the style consistency subset of the page is defined as $ S \subseteq W, \forall e_i, e_j \in S, style(e_i) = style(e_j).$

Thus the real web page \(W\) and its generated page \(\hat{W}\) are divided into style consistency subsets \(W = \{S_j \mid j \in M\}\) and \(\hat{W} = \{S_k \mid k \in N\}\), respectively. Given $N$ and $M$ can differ, we apply a \texttt{max} operation for optimal matching. The SC Score is then calculated as:
\begin{equation}
     SCScore(W, \hat{W}) = \sum_{j=1}^{M} w_j \cdot \max_{k} J(S_j, \hat{S}_k),
\label{eq:sc_score}
\end{equation}
where \( J(A, B) \) is the Jaccard similarity coefficient. Additionally, under the assumption that style consistency subsets with more elements are more semantically valuable, we utilize a weight $w_j = \frac{|S_{j}|}{\sum_{l=1}^{M}|S_{l}|}$.

\subsection{Implementation}
\label{sec:implementation}
Two baselines are implemented: autoregressive (\textbf{WebRPG-AR}) and diffusion model (\textbf{WebRPG-DM}). The VAE, hierarchical embedding, and character count embedding are jointly trained with the backbone, and the semantic embedding is produced by frozen pre-trained MarkupLM$_{large}$ \cite{Li2021MarkupLMPO}. The XPath embedding layer is initialized following  Li et al. \cite{Li2021MarkupLMPO}.  All baselines are based on Transformer \cite{vaswani2017attention} and have approximately 50M of parameters to ensure fair comparison, whose hidden dimensions are 128. The dimensions of latent vector and HTML embedding $d$ is 128. For optimization, AdamW \cite{adamw} is used with a learning rate of 1.2e-4. All models are trained for 1M steps with a batch size of 300. 

Additionally, LLMs have been gaining adoption in different domains. We assess GPT-4 \cite{yang2023dawn,openai2023gpt4}, StarCoder2-7b \cite{lozhkov2024starcoder}, DeepSeek-Coder-6.7b \cite{guo2024deepseek}, CodeLlama-13B \cite{roziere2023code} on the WebRPG task. GPT-4 is one of the state-of-the-art LLMs, while the others are open-source models known for code generation. Due to limited resources, we randomly select 10\% of test samples. The prompt template employs in-context learning \cite{GPT3}, incorporating a task description, three demonstrations, and a test instance. Further details are available in Sec. A.5.

\subsection{Quantitative and Qualitative Evaluation}
\label{sec:experiment_result}
\begin{figure}[t]
\centering
\includegraphics[width=0.99\textwidth]{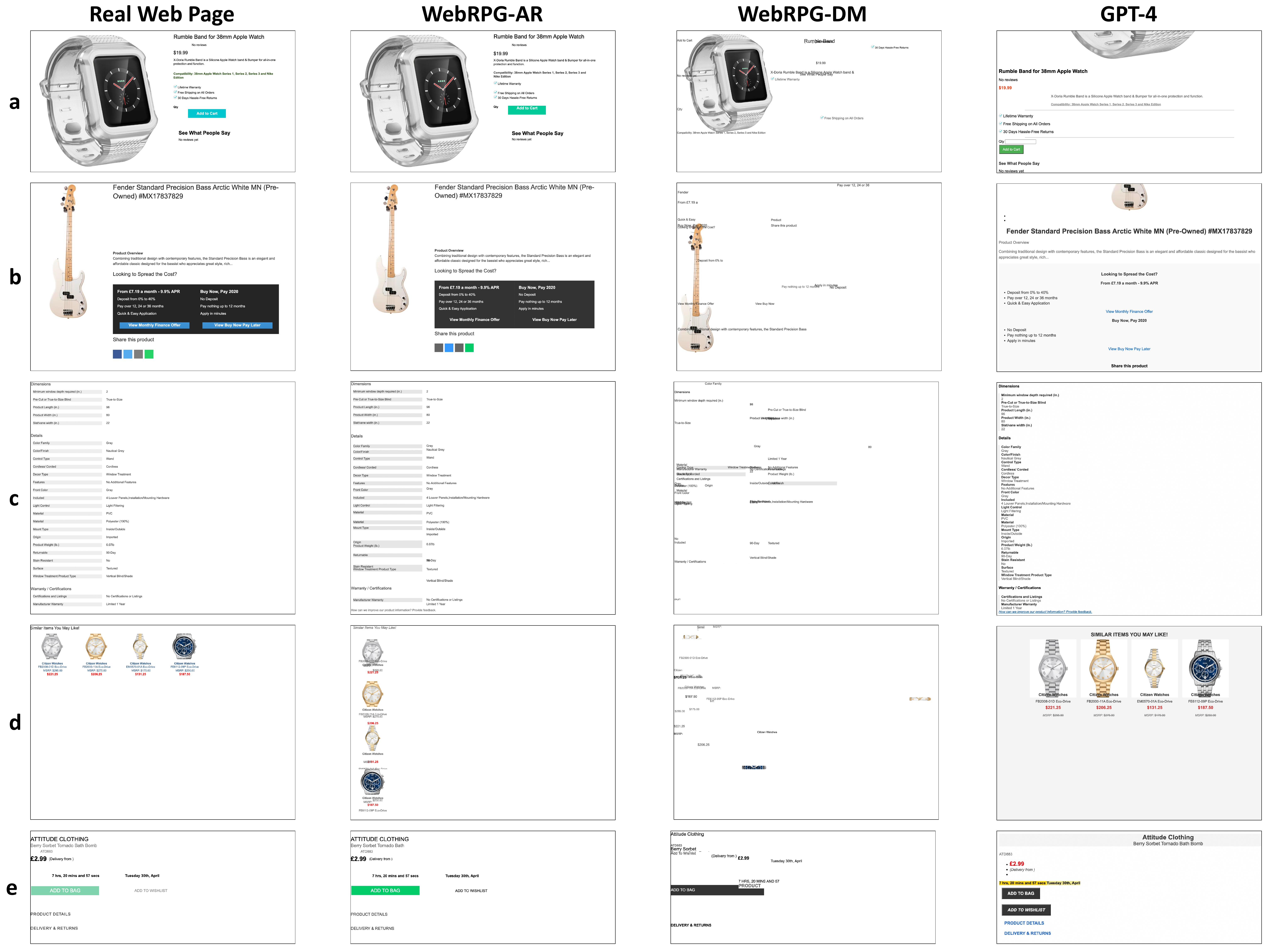} 
\caption{Qualitative comparison of WebRPG baselines.}
\label{fig:cases}
\end{figure}

We present quantitative results in \cref{tab:results} and qualitative results in \cref{fig:cases}. Regarding the results of real data, in addition to the normally rendered web page (\cref{tab:results}, ``Real Web Pgae''), we also report the web page rendered using only HTML code (\cref{tab:results}, ``Plain HTML''). Since the browser would apply default CSS when custom CSS is absent, some models perform worser than the plain HTML due to unreasonable generated RPs. The FIDs for real data are calculated between the test set and other real web pages.

The experimental results show that WebRPG-AR consistently surpasses other baselines. Its sequential decoding mechanism allows for more refined control based on previously generated results \cite{Du2023ContextPP,Xiao2022ASO}. As shown in \cref{fig:cases} a, b, e, WebRPG-AR demonstrates impressive visual quality in detail.

\begin{table}[]
\caption{WebRPG baselines quantitative comparison with bold figures for best results. "*" stands for the result in the randomly selected test set.}
\centering
\setlength\tabcolsep{5pt}
\begin{tabular}{lccccc}
\toprule
                     & Overall          & \multicolumn{2}{c}{Layout}                                 & \multicolumn{2}{c}{Style}                                 \\
Model                & FID $\downarrow$ & FID$_{\rm{layout}}$    $\downarrow$ & Ele. IoU  $\uparrow$ & FID$_{\rm{style}}$    $\downarrow$ & SC Score  $\uparrow$ \\
\midrule
WebRPG-AR            & \textbf{0.1281}  & \textbf{0.1520}                     & \textbf{0.7082}      & \textbf{0.2124}                    & \textbf{0.9474}      \\
WebRPG-DM            & 62.021           & 60.942                              & 0.0357               & 106.95                             & 0.3671               \\
\midrule
WebRPG-AR*           & 0.1324           & 0.2877                              & 0.7069               & 0.1359                             & 0.9485               \\
WebRPG-DM*           & 61.135           & 60.870                              & 0.0356               & 105.86                             & 0.3649               \\
GPT4*                & 4.2141           & 47.732                              & 0.0347               & 8.8898                             & 0.5515               \\
StarCoder2-7b*       & 11.899           & 51.432                              & 0.0309               & 18.186                             & 0.3639               \\
DeepSeek-Coder-6.7b* & 5.8219           & 55.744                              & 0.0330               & 7.4542                             & 0.3949               \\
CodeLlama-13b*       & 9.2826           & 55.427                              & 0.0278               & 11.625                             & 0.3864               \\
\midrule
Real Web Page            & 0.0027           & 0.0015                              & 1.0000               & 0.0074                             & 1.0000               \\
Plain HTML           & 8.5342           & 52.438                              & 0.0354               & 8.4951                             & 0.3668              \\
\bottomrule
\end{tabular}
\label{tab:results}
\end{table}

\begin{table}[tb]
\caption{Ablation study based on WebRPG-AR. Best results in bold. ``$\mathcal{Z}_{mask}$'' is detailed in \cref{sec:core_model}. ``H.E.'' stands for HTML embedding. ``S.'', ``H.'', and ``C.'' stand for semantic, hierarchical, and character count embeddings.}
\centering
\setlength\tabcolsep{3.2pt}
\begin{tabular}{lccccclccccc}
\toprule
   &            &                      &        \multicolumn{3}{c}{H. E.}    &  & Overall          & Layout                              &                      & Style                              &                      \\
\# & VAE        & $\mathcal{Z}_{mask}$ & S.         & H.         & C.         &  & FID $\downarrow$ & FID$_{\rm{layout}}$    $\downarrow$ & Ele. IoU  $\uparrow$ & FID$_{\rm{style}}$    $\downarrow$ & SC Score  $\uparrow$ \\
\midrule
1  &            & \checkmark           & \checkmark & \checkmark & \checkmark &  & 0.9702           & 5.4668                              & 0.5954               & 15.923                             & 0.8053               \\
2  & \checkmark &                      & \checkmark & \checkmark & \checkmark &  & 0.1487           & 0.2055                              & 0.6462               & 0.2944                             & 0.9332               \\
\midrule
3  & \checkmark & \checkmark           &            & \checkmark & \checkmark &  & 0.1797           & 0.2096                              & 0.6620               & 0.3055                             & 0.9323               \\
4  & \checkmark & \checkmark           & \checkmark &            & \checkmark &  & 0.3003           & 0.3770                              & 0.6345               & 1.9048                             & 0.8982               \\
5  & \checkmark & \checkmark           & \checkmark & \checkmark &            &  & 0.1575           & 0.3152                              & 0.6769               & 0.3065                             & 0.9434               \\
\midrule
6  & \checkmark & \checkmark           & \checkmark & \checkmark & \checkmark &  & \textbf{0.1281}  & \textbf{0.1520}                     & \textbf{0.7082}      & \textbf{0.2124}                    & \textbf{0.9474}     \\
\bottomrule
\end{tabular}
\label{tab:ablation}
\end{table}

The performance of WebRPG-DM is suboptimal across all metrics. It only tends to produce standard web visual presentations in simpler cases, as illustrated in \cref{fig:cases} e, such as bolding prices, adding background color to buttons, and aligning a few elements.  This implies that diffusion models may be inappropriate for this task. There are two plausible explanations: First, unlike images and videos in Euclidean space, web elements are non-Euclidean due to their hierarchical arrangement, while diffusion models are confined to Euclidean space \cite{Koo2023ASO}. Second, the WebRPG task demands meticulous adjustments and detailed control for realism, a limitation of diffusion models~\cite{diffusion-lm}. 

GPT-4's performance on the WebRPG task surpasses that of WebRPG-DM and falls short of WebRPG-AR. Open-source LLMs underperform compared to GPT-4. As illustrated in the \cref{fig:cases} a,b,e, GPT-4 can effectively handle element styles, such as adding background colors to buttons and applying distinct colors for prices. However, the performance of GPT-4 in layout is limited. As demonstrated in \cref{fig:cases} a-c, GPT-4 tends to generate simplistic vertical arrangements when faced with complex HTML structures. With regular HTML, as depicted in \cref{fig:cases} e, GPT-4 achieves a layout that is similar to the real page. Therefore, we conclude that GPT-4 demonstrates basic capability in WebRPG tasks with regular HTML, but its performance with complex HTML is less effective. Additionally, we notice that LLMs do not generate RPs for all elements, causing many to use the browser's default CSS, resulting in performance similar to plain HTML.

It is worth noting that WebRPG-AR exhibits the ability to render diverse web pages. For example, \cref{fig:cases} d shows WebRPG-AR's creation of a page with a vertical layout (originally horizontal), preserving the pattern and order consistency across four groups. This finding suggests that the model successfully learns web design knowledge and applies it effectively to render web pages from HTML code. Further cases are available in Sec. B.1.

Furthermore, we calculate the FID on screenshots of rendered web pages, following conventional image generation practices \cite{ramesh2022hierarchical,stable_diffusion}. The results, shown in Sec. B.2, are consistent with \cref{tab:results}. Additionally, we conduct a human evaluation, detailed in Sec. B.4, with results that also align with \cref{tab:results}.

\begin{figure}[t]
\centering
\includegraphics[width=0.95\columnwidth]{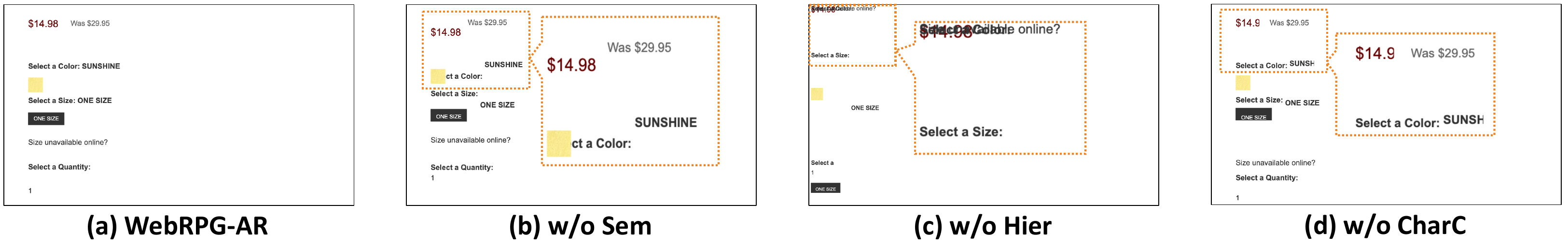} 
\caption{ Case visualization from the ablation study. }
\label{fig:ablation}
\end{figure}

\subsection{Ablation Study}
We conduct a series of ablation experiments based on WebRPG-AR, as shown in \cref{tab:ablation}. \#1 uses a one-dimensional flat input instead of VAE. \#2 removes $\mathcal{Z}_{mask}$ (in \cref{sec:core_model}). \#3 and \#5 respectively remove the corresponding embedding layers, while \#4 substitutes hierarchical embedding with one-dimensional positional embedding. Additionally, we visualize some cases from \#3, \#4, and \#5 in \cref{fig:ablation}. All models are trained to convergence following the settings in \cref{sec:implementation}.

The results of \#1 demonstrate the effectiveness of using VAE for rendering parameters compression. Although \#2 is comparable to \#6, the incorporation of \(\mathcal{Z}_{mask}\) enhances the model stability during training.
The results of \#3, \#4, and \#5 reveal that all three embeddings play critical roles in web design. Hierarchical embedding helps layout arrangement significantly. The simplification to 1D positional embedding leads to a disorganized layout, as illustrated in \cref{fig:ablation} c. Semantic embedding enhances the model with the capacity to perceive semantic relationship. For example, as \cref{fig:ablation} b shows, the model struggles to horizontally align elements like ``select a color'' and ``sunshine,'' suggesting challenges in identifying key-value pairs without semantic information. Character count embedding helps to predict appropriate element sizes for full content display, as in \cref{fig:ablation} d, where a narrow ``price'' and ``sunshine'' width leads to incomplete text display.

\begin{figure}[t]
\centering
\includegraphics[width=0.99\columnwidth]{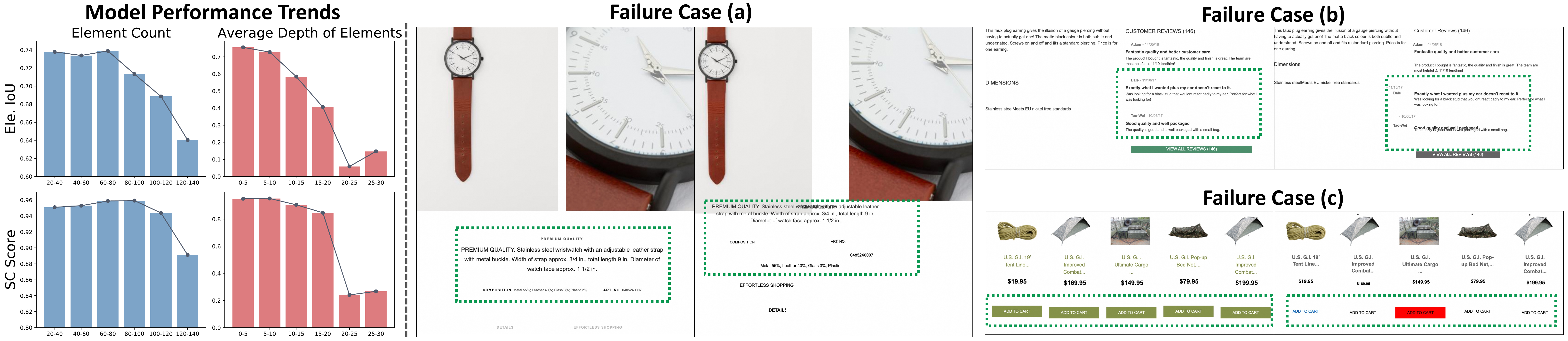} 
\caption{ \textit{Left}: Trends in WebRPG-AR performance relative to the number of elements and average depth of elements within the DOM tree. \textit{Right}: WebRPG-AR failure cases with real web pages on the left, generated results on the right, and highlights in \textcolor[RGB]{13, 154, 83}{\textbf{green}}.
}
\label{fig:failure_cases}
\end{figure}

\subsection{Discussion on Failure Cases}

To investigate the boundaries of the model's capabilities, we analyze several failure cases generated by WebRPG-AR. The left side of \cref{fig:failure_cases} reveals that both layout (Ele. IoU) and style (SC Score) metrics decrease with an increase in the number of elements or the average depth of elements within the DOM tree. This trend may be attributed to two factors: the inherent complexity of a page increases with more elements or greater depth, and the training set lacks web pages with a large number of elements or significant depths (details in Sec. A.3). Regarding error types, layout issues mainly include misalignments and overlaps, as shown in \cref{fig:failure_cases} a and b. For style, the model struggles to recognize web page elements with identical semantic functions, such as the ``Add to Cart'' buttons illustrated in \cref{fig:failure_cases} c, which should appear identical. Moreover, we observe two primary error scenarios: elements positioned at the end of the HTML code tend to be more error-prone, as seen with the element in the bottom right corner of \cref{fig:failure_cases} b, likely due to the characteristics of the autoregressive model \cite{Dong2023ASO}; additionally, pages with large-scale images pose challenges, as shown in \cref{fig:failure_cases} a, since the model does not take the original images as input. The discussion above highlights the need for further research.

\begin{figure}[t]
\centering
\includegraphics[width=0.95\columnwidth]{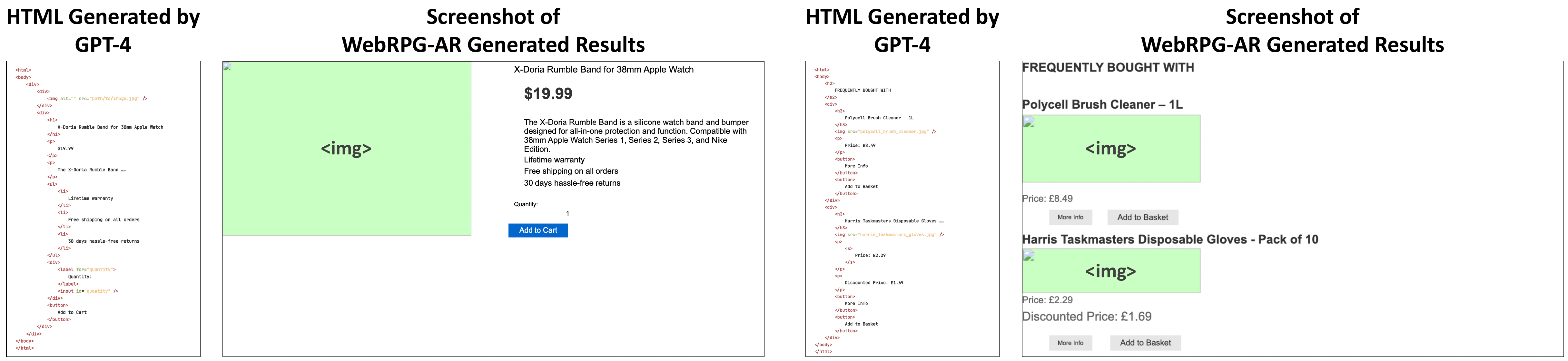} 
\caption{ The HTML code generated by GPT-4 and the corresponding web page visual results generated by WebRPG-AR. Screenshots use \textcolor[RGB]{13, 154, 83}{\textbf{green}} \(\textless\)img\(\textgreater\) placeholders. }
\label{fig:GPT_Gen_HTML_small}
\end{figure}

\subsection{Discussion on the Integration of LLM and WebRPG Model}
Recently, LLMs have enabled the possibility of automatically generating HTML code \cite{li2023enabling}. Consequently, we hypothesize that integrating LLM into a WebRPG system could facilitate a fully automated web development workflow. We employ GPT-4 \cite{yang2023dawn,openai2023gpt4} to validate this hypothesis. As \cref{fig:GPT_Gen_HTML_small} illustrates, WebRPG-AR effectively creates visual presentations of web pages based on generated HTML, demonstrating the potential of a fully automated web development workflow through the integration of LLM and WebRPG. Additional cases and the prompt for automatically generating HTML are provided in Sec. B.3.

%% file: 6_Conclusion.tex
\section{Conclusion and Limitations}
This paper presents WebRPG, a task that automates web design by generating rendering parameters for web elements from HTML. We introduce a new dataset, two baseline models, and evaluation metrics. Results show the autoregressive baseline most effectively generates web visual presentations.

Nevertheless, this study has limitations that warrant further investigation in future research.
The proposed model can undergo fine-tuning to support design tasks such as partial web page design by masking specific elements. Additionally, it can be adapted to analyze raster images by replacing \(\textless\)img\(\textgreater\) tokens with image embeddings.
The employment of established CSS frameworks like Tailwind\footnote{\url{https://tailwindcss.com/}} could standardize CSS, thereby potentially simplifying the WebRPG task. However, sourcing web pages based on these frameworks presents challenges.
Furthermore, design options and control mechanisms of the results are worth exploring.
Future research will address these aspects.

%% file: X_suppl.tex
\clearpage
\setcounter{page}{1}
\renewcommand\thesection{\Alph{section}}
\setcounter{section}{0}

\begin{center}
\LARGE\textbf{Supplementary Material}
\end{center}

\section{Additional Details}
\subsection{Details of Rendering Parameters}
\label{sec:details_rps}
\begin{figure}[h]
\centering
\includegraphics[width=0.85\columnwidth]{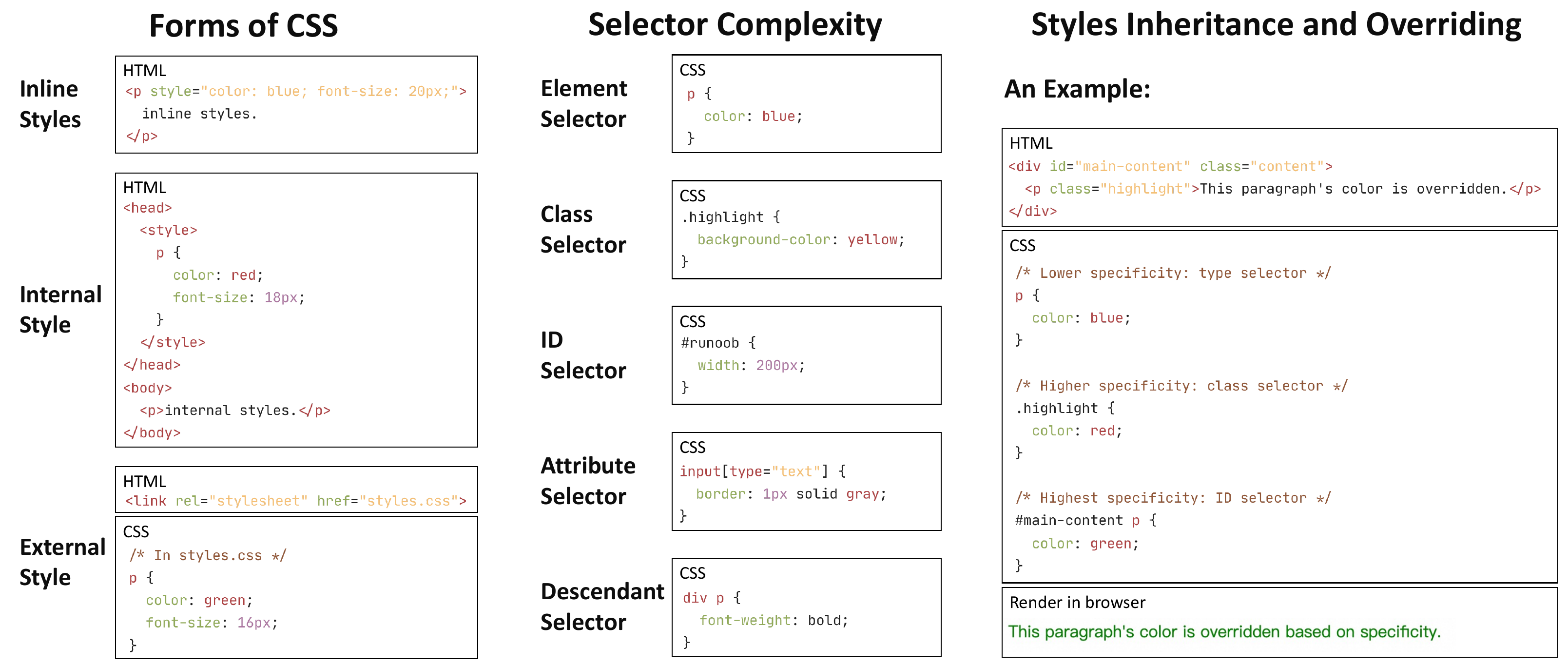} 
\caption{Examples of CSS code complexity, showcasing various CSS forms (\textit{left}), selector complexity (\textit{middle}), and style inheritance and overrides (\textit{right}).}
\label{fig:css_complexity}
\end{figure}

As described in \cref{sec:task_fotmula}, we utilize rendering parameters to standardize CSS due to its code complexity. The examples in \cref{fig:css_complexity} demonstrate this complexity. As shown on the left side of \cref{fig:css_complexity}, CSS can be utilized in different forms\footnote{\url{https://www.w3schools.com/css/css_howto.asp}}: \texttt{Inline Styles} for direct HTML element styling via the ``style'' attribute;\texttt{ Internal Style Sheets}  using ``\(\textless\)style\(\textgreater\)'' tags within HTML documents; and \texttt{External Style Sheets} linking to CSS files externally. The middle of \cref{fig:css_complexity} showcases various CSS selectors\footnote{\url{https://www.w3schools.com/css/css_selectors.asp}}, including simple \texttt{tag}, \texttt{class}, and \texttt{ID} selectors, as well as complex \texttt{attribute} and \texttt{descendant} selectors. Furthermore, CSS follows certain rules regarding inheritance and overrides\footnote{\url{https://developer.mozilla.org/en-US/docs/Web/CSS/Inheritance}}. An example on the right side of \cref{fig:css_complexity} shows how the \texttt{.highlight} class's red color is overridden by the more specific ID selector \texttt{\#main-content} p, turning the color green.

The complexity of CSS makes direct generation of CSS impractical. Even parsing CSS code to obtain WebRPG task labels is challenging. Since browsers compute the final applied CSS property values (i.e., rendering parameters) for each element based on HTML and CSS to render web pages, we propose extracting each element's RPs directly from the browser, as described in \cref{sec:rp_definition}. This approach bypasses the need to parse CSS code, achieving the standardization of CSS.

\begin{figure}[h]
\centering
\includegraphics[width=0.85\columnwidth]{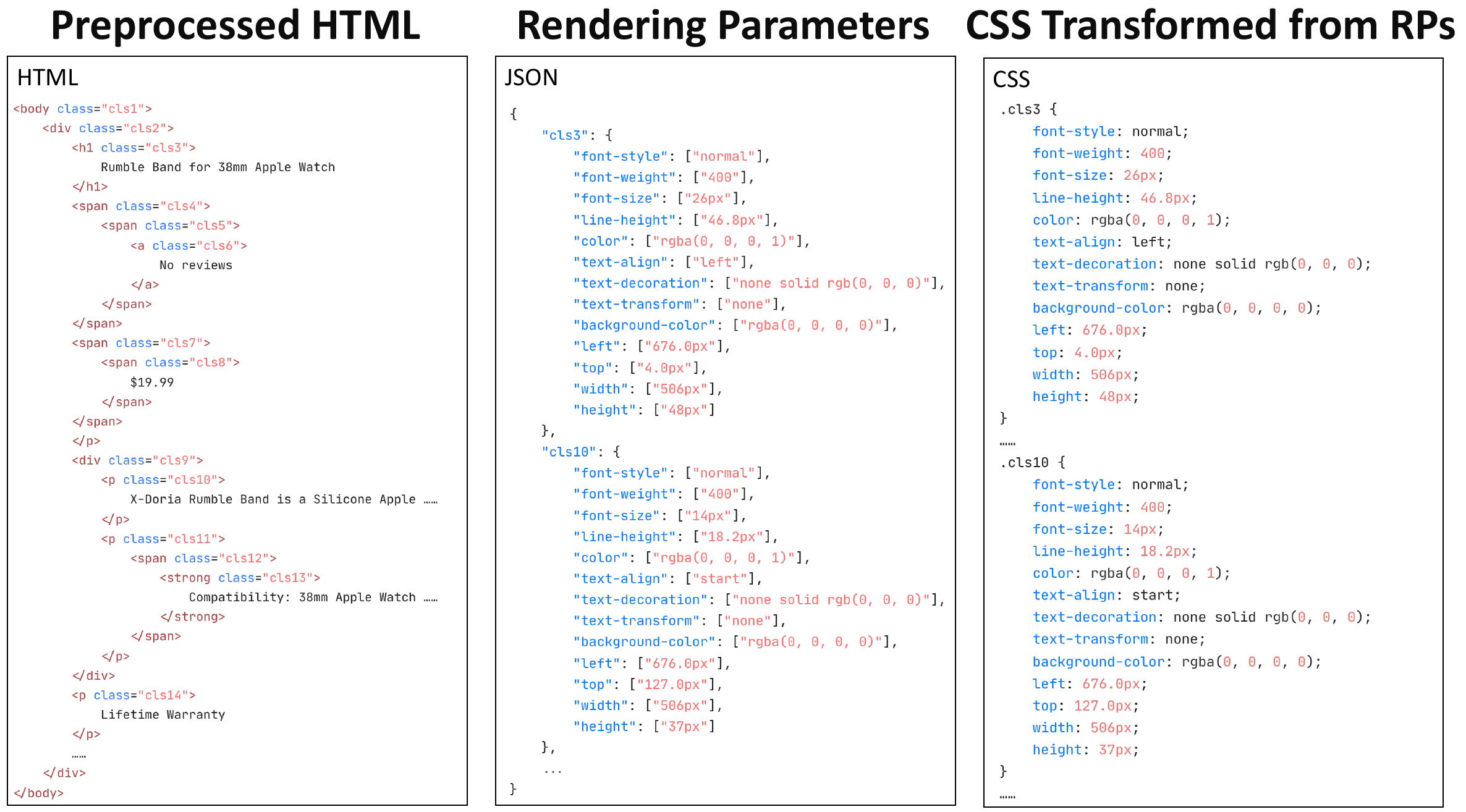} 
\caption{A illustration case of rendering parameters organization, including preprocessed HTML (\textit{left}), JSON-stored rendering parameters (\textit{middle}), and the CSS  transformed from those RPs (\textit{right}).}
\label{fig:rps_css}
\end{figure}

\begin{table}[h]
\centering
\setlength\tabcolsep{3.2pt}
\caption{ The complete vocabulary of rendering parameters including all categories, their index ranges, and selected examples.}
\centering
\begin{tabular}{llc}
\toprule
Category        & Index Range     & Examples                                    \\
\midrule
Integer Pixel   & 0-1920    & 1px, 1052px, 1920px                         \\
Color           & 1921-1966 & RGBA(153, 204, 0, 1), RGBA (255, 255, 255, 1) \\
Font Style      & 1967-1969 & italic, oblique                             \\
Font Weight     & 1970-1978 & 100, 500, 900                                 \\
Line Height     & 1979      & normal                                      \\
Text Align      & 1980-1985 & start, center, end                          \\
Text Decoration & 1986-1987 & none, underline                             \\
Text Transform  & 1988-1991 & uppercase, capitalize                       \\
PAD             & 1992      & PAD                                         \\
\bottomrule
\end{tabular}
\label{tab:vocabulary}
\end{table}

\begin{table}[h]
\centering
\setlength\tabcolsep{15pt}
\caption{ Index ranges for each rendering parameter in the vocabulary.}
\begin{tabular}{ll}
\toprule
Rendering Parameter       & Index Range \\
\midrule
\textit{left}             & 0-1920      \\
\textit{top}              & 0-1920      \\
\textit{width}            & 0-1920      \\
\textit{height}           & 0-1920      \\
\textit{font-style}       & 1967-1969   \\
\textit{font-weight}      & 1970-1978   \\
\textit{font-size}        & 0-32        \\
\textit{line-height}      & 0-50, 1979  \\
\textit{text-align}       & 1980-1985   \\
\textit{text-decoration}  & 1986-1987   \\
\textit{text-transform}   & 1988-1991   \\
\textit{color}            & 1921-1966   \\
\textit{background-color} & 1921-1966   \\
\bottomrule
\end{tabular}
\label{tab:index_range}
\end{table}

In practice, we follow the pre-order traversal order of the DOM tree to assign a unique ID to each element, achieved by modifying the class name, as shown on the left side of \cref{fig:rps_css}. We organize the rendering parameters using JSON, where the key is the element's ID, as illustrated in the middle of \cref{fig:rps_css}. RPs can also be transformed into CSS, utilizing class selectors only, as demonstrated on the right side of \cref{fig:rps_css}.

Additionally, the complete vocabulary of all rendering parameters is detailed in \cref{tab:vocabulary}, and index ranges of each rendering parameter are presented in \cref{tab:index_range}.

\subsection{Details of Visual Complexity Metric}
\label{sec:VC}
The Visual Complexity (VC) metric integrates three dimensions: color, size, and alignment. For any given web page, the three dimensions are defined as follows:

\textbf{Color:} The color metric measures the richness of colors and is defined as:
\begin{equation}
  VC_{color} = \frac{1}{2N}(C_c + C_{bg} - 2),
  \label{eq:vc_color}
\end{equation}
where $N$ is the number of elements, and $C_c$ and $C_{bg}$ are the counts of unique  \textit{color} and \textit{background-color} attributes respectively.

\textbf{Size:} The size metric measures the diversity of sizes among web page elements. In particular, it calculates the size diversity for all \(N'\) parent elements and then computes the average. The formula is as follows:
\begin{equation}
   VC_{size} =  \frac{1}{{N'}}{\sum_{i=1}^{N'}\left(\frac{DS_i-1}{NC_i}\right)},
  \label{eq:vc_size}
\end{equation}
with $NC_i$ and $DS_i$ being the count of child elements and their distinct sizes for element $i$, respectively.

\textbf{Alignment:} The complexity of a web page inversely correlates with the number of pairwise alignments \cite{fu2007measuring}. To simplify, this metric applies only to leaf nodes.  The calculation formula is as follows:

\begin{equation}
   VC_{alg} = 1 - \frac{1}{N_{leaf}(N_{leaf}-1)}\sum_{j=1}^{N_{leaf}}\sum_{i \neq j}^{N_{leaf}} ALG_{ij},
  \label{eq:vc_alignment}
\end{equation}
where $N_{leaf}$ denotes the number of leaf node elements, and $ALG_{ij}$ is a binary indicator of alignment (1) or misalignment (0) between elements $i$ and $j$.

The overall VC is the sum of three metrics: \(VC = VC_{\text{color}} + VC_{\text{alg}} + VC_{\text{size}}\).

\subsection{Dataset Details}
\label{sec:details_dataset}

\begin{figure}[h]
\centering
\includegraphics[width=0.55\columnwidth]{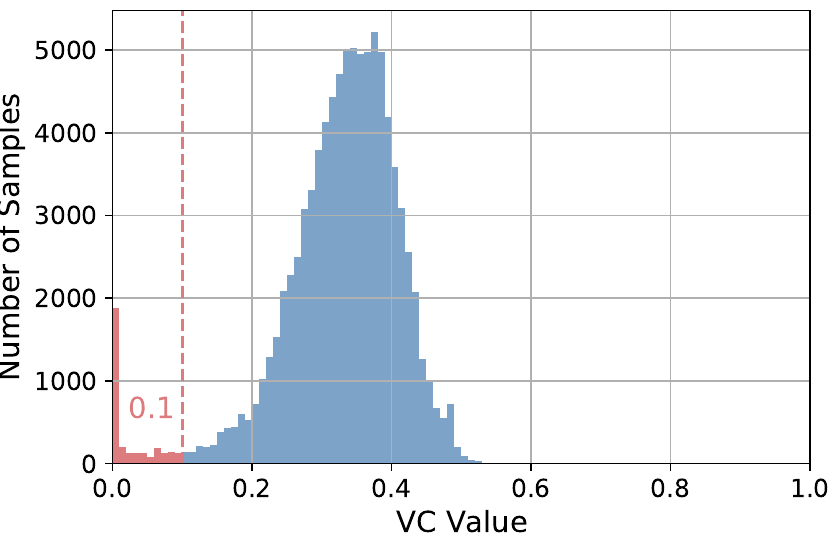} 
\caption{Histogram showcasing Visual Complexity (\cref{sec:VC}) value distribution across all samples. Red indicates samples are filtered out, while blue represents those retained in the dataset.}
\label{fig:vc_distribution}
\end{figure}

\begin{figure}[h]
\centering
\includegraphics[width=0.95\columnwidth]{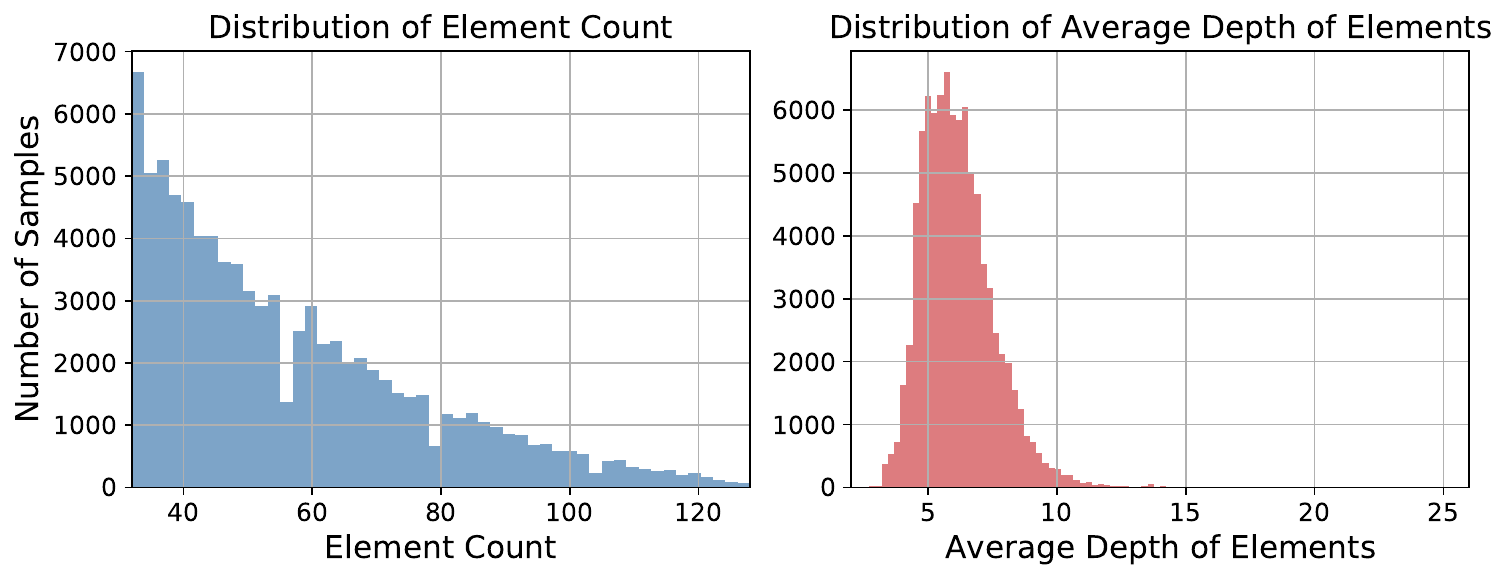} 
\caption{Histograms showcasing element count and the average depth of elements distribution across all samples in the dataset.}
\label{fig:dataset_histogram}
\end{figure}

The distribution of Visual Complexity (\cref{sec:VC}) values across all samples is illustrated in \cref{fig:vc_distribution}. In our dataset, samples with a VC value of less than 0.1 are filtered out, resulting in a remaining subset where the VC distribution is relatively concentrated and approximates a normal distribution, thereby helping to mitigate the impact of extreme samples on training. Additionally, to further investigate our dataset, we visualize two crucial statistical values, element count and the average depth of elements, in \cref{fig:dataset_histogram}. This visualization indicates that the dataset lacks samples containing a large number of elements or considerable element depths.

\subsection{Implementation details of FID model}
\label{sec:details_fid_model}
As described in \cref{sec:fid}, the FID model is a binary classifier, incorporating a VAE described in \cref{sec:vae}, four transformer layers, and a classification header. A special \textit{CLS} vector is utilized as the classification feature, representing all RPs. The rest of the input is the same as the model in \cref{sec:mle_model}. Three kinds of noise are designed to pollute the real data, namely perturbing the original values with a fixed variance, randomly substituting elements with synthetic ones, and randomly swapping elements.  The specific FID models for layout and style, namely FID\(_{layout} \) and FID\(_{style} \), are trained by masking irrelevant inputs. Specifically, FID\(_{layout} \) processes only the layout, masking the style, and FID\(_{style} \) processes only the style, masking the layout. The FID models for overall, layout, and style, achieve classification accuracies of 88.8\%, 95.5\%, and 92.4\%, respectively.

\begin{table}[t]
\centering
\setlength\tabcolsep{5pt}
\caption{The prompt template for GPT-4 experiment in \cref{sec:experiment_result}.}
\begin{tabular}{llp{7cm}}
\toprule
Prompt &
  \multicolumn{2}{p{9.5cm}}{You are an exceptional web designer. Please create the corresponding CSS code based on the HTML code I have provided, so as to craft a well-designed visual presentation for the web page. You can only use the following CSS properties:   "left", "top", "width", "height",   "font-style", "font-weight", "font-size", "line-height", "color", "text-align",   "text-decoration", "text-transform", "background-color".   Please exercise caution in controlling the size of the image, as using the original image dimensions directly may result in excessive spatial occupation. Here are several demonstrations:\{Demonstrates\}. Below is the HTML code and do not reply with anything other than CSS code: \{HTML\_Code\}}. \\
\midrule
Slots &
  Demonstrates &
  The HTML-CSS pairs for three selected web page segments. \\
 &
  HTML\_Code &
  HTML code of given web page. \\
\bottomrule
\end{tabular}
\label{tab:gpt_4_prompt}
\end{table}

\subsection{Implementation details of WebRPG Baselines}
\label{sec:implement_details}

The backbone of WebRPG-AR consists of 6-layer transformers for both encoder and decoder, and WebRPG-DM is a 12-layer U-ViT. The mask scheduling function $\gamma(r)$ is a cosine function, the time steps $T$ in diffusion follows \cite{ddpm} with a value of 1000, and $\lambda_{KL}$ is set to 1e-6. For optimization, AdamW \cite{adamw} is used with a learning rate of 1.2e-4, $\beta_1$ of 0.9, and $\beta_2$ of 0.99.

The prompt template for the LLMs experiment in \cref{sec:experiment_result} is detailed in \cref{tab:gpt_4_prompt}. Due to the extensive length of textual representation for each element's RPs, as shown on the right side of \cref{fig:rps_css}, we opt to have LLMs directly generate the CSS code. The specific steps for conducting the LLMs experiment are:

\begin{enumerate}
    \item Use the prompt to generate CSS code via LLMs.
    \item Use a browser to render the web page with the given HTML and the CSS code generated by LLMs.
    \item Extract the RPs for all elements, employing the method in \cref{sec:data_pre_process}.
    \item Evaluate these RPs using the metrics in \cref{sec:metric}.
\end{enumerate}

\begin{figure}[h]
\centering
\includegraphics[width=0.99\textwidth]{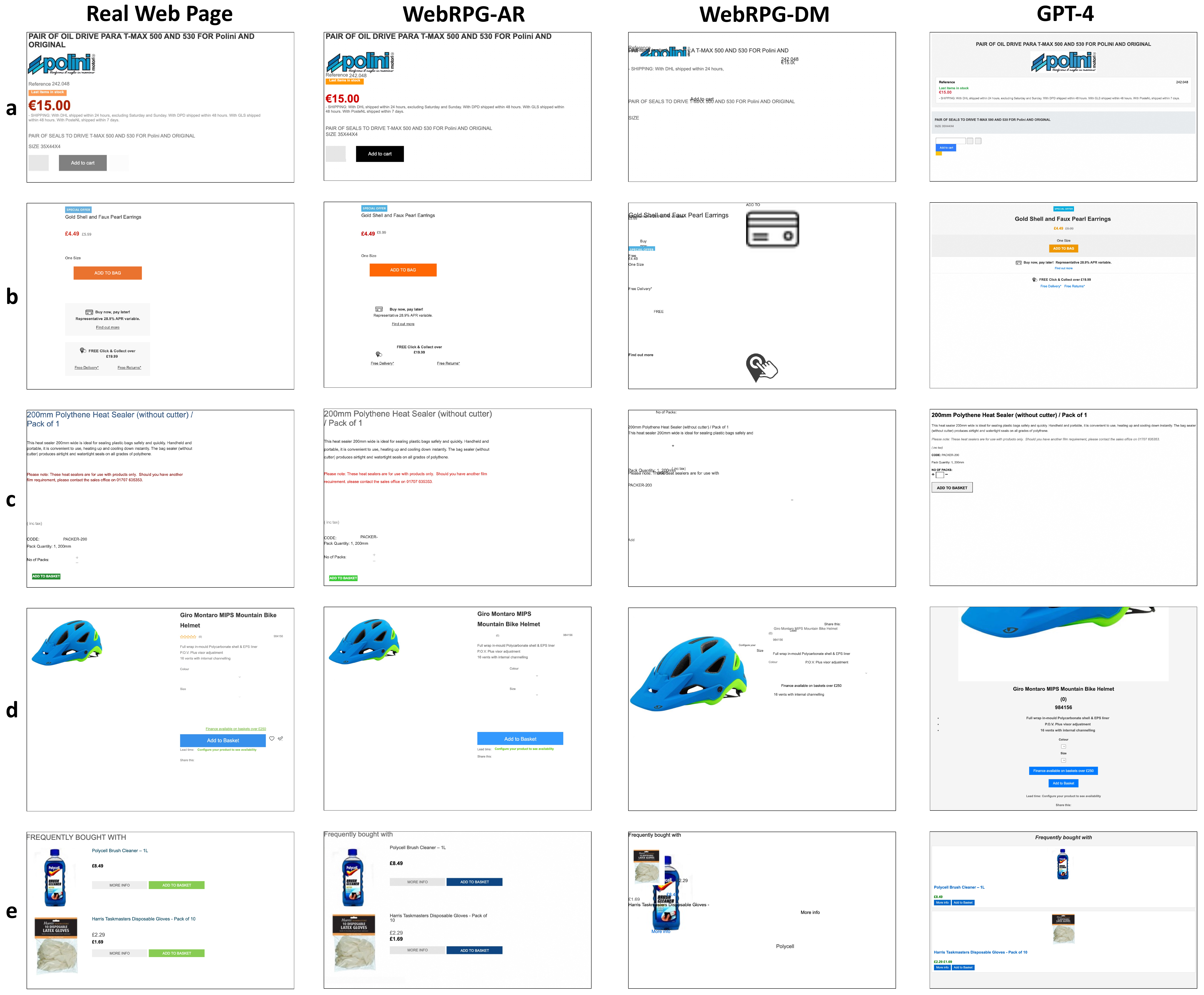} 
\caption{Additional visualization of baseline-generated results. The screenshots focus on areas with elements.}
\label{fig:additional_cases}
\end{figure}

\section{Additional Results}

\subsection{Additional Cases of Baseline-Generated Results}
\label{sec:cases_baseline}
\begin{figure}[h]
\centering
\includegraphics[width=0.99\textwidth]{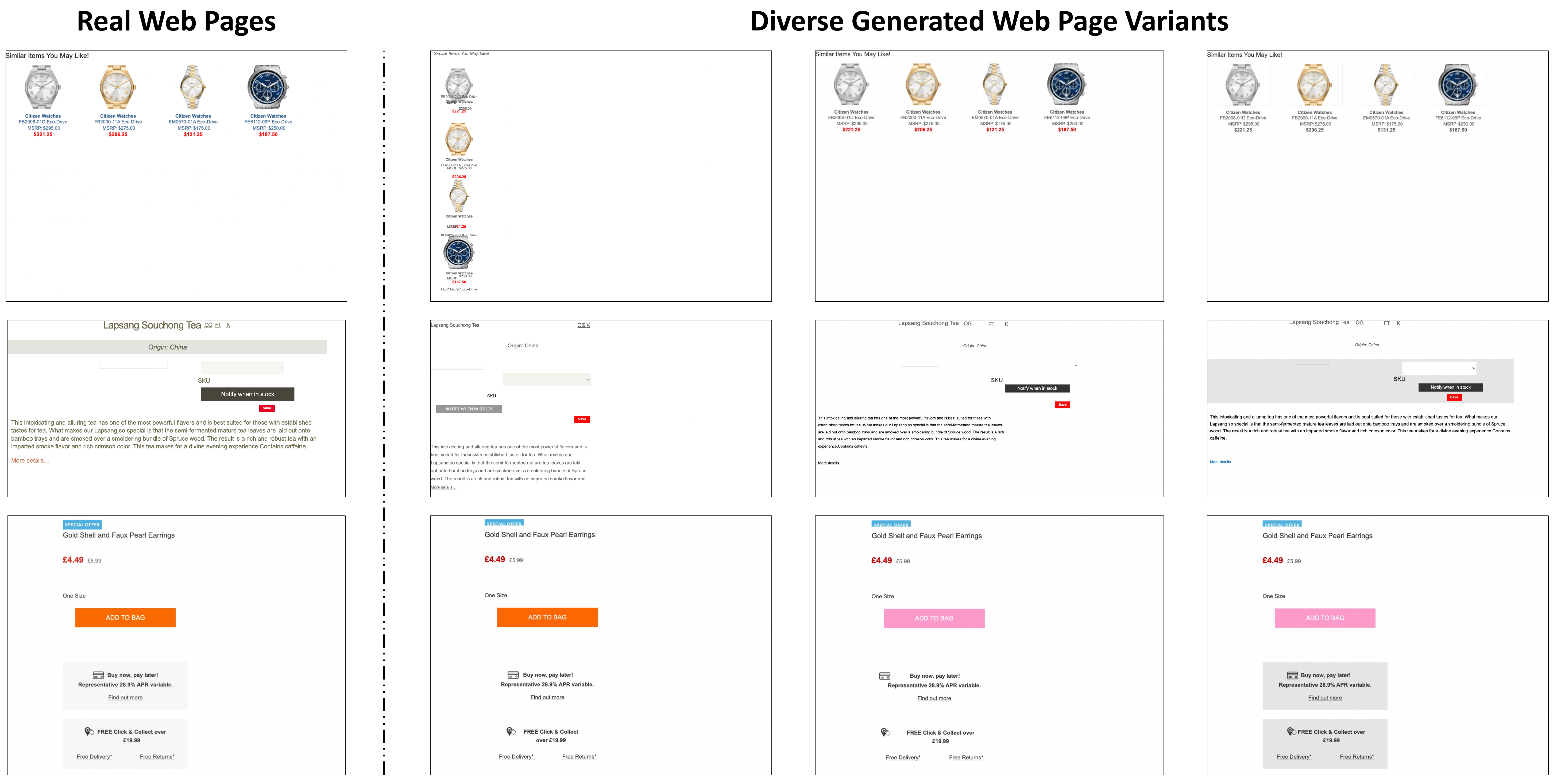} 
\caption{The web page variants generated by WebRPG-AR based on the same HTML.}
\label{fig:diversity}
\end{figure}

We present additional results from WebRPG baselines in \cref{fig:additional_cases}. These results exhibit the performance of all baselines comparable to that outlined in \cref{sec:experiment_result}.
Additionally, \cref{fig:diversity} displays the web page variants generated by WebRPG-AR based on the same HTML, each produced through individual inferences. The differences in layout and style among these variants indicate that WebRPG-AR can generate diverse web pages while maintaining semantic coherence.

\begin{figure}[h]
\centering
\includegraphics[width=0.95\textwidth]{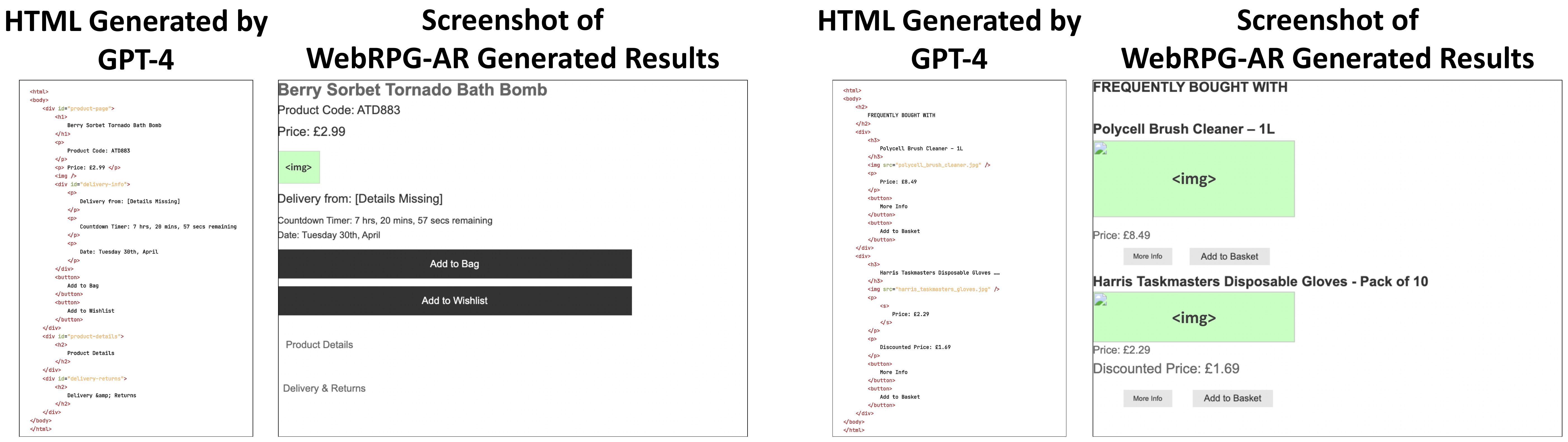} 
\caption{The HTML code generated by GPT-4 and the corresponding web page visual results generated by WebRPG-AR. Screenshots use \textcolor[RGB]{13, 154, 83}{\textbf{green}} \(\textless\)img\(\textgreater\) placeholders due to GPT-4 generates fictitious source addresses.}
\label{fig:GPT_Gen_HTML}
\end{figure}

\begin{table}[h]
\setlength\tabcolsep{9pt}
\caption{FID on rendered web page screenshots.} \label{tab:FID_screenshot}
\begin{tabular}{lcccc}
\toprule
                        & WebRPG-AR     & GPT4   & WebRPG-DM     & Real Web Page \\
\midrule
FID$_{\rm{Screenshot}}$ & 3.2102 & 15.515 & 33.040 & 1.1156   \\ 
\bottomrule
\end{tabular}
\end{table}

\subsection{The FID on Screenshots of Rendered Web Pages}
\label{sec:fid_screenshot}
The FID on screenshots of rendered web pages is shown in \cref{tab:FID_screenshot}.

\subsection{Further Cases of Integrating LLM with WebRPG Model}
\label{sec:integrate_llm}
\cref{fig:GPT_Gen_HTML} showcases more cases of WebRPG-AR creating visual presentations of web pages based on HTML code generated by GPT-4. The prompt template for automatically generating HTML is in \cref{tab:template_HTML}. The prompt encompasses human-authored descriptions of web design ideas, with an example shown in \cref{tab:design_idea}.

\begin{figure}[t]
  \centering
    \includegraphics[width=1\linewidth]{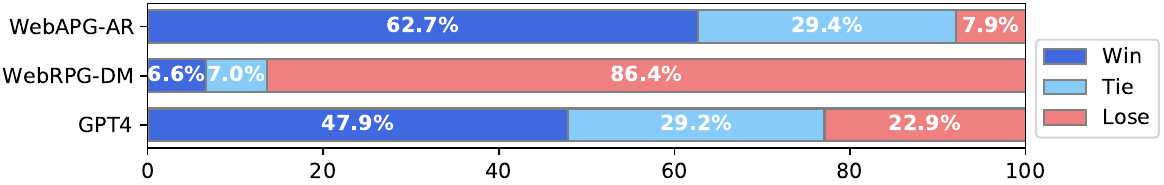}
   \caption{Human pairwise comparison evaluation results.}
   \label{fig:human_eval}
\end{figure}

\begin{table}[h]
\caption{The prompt template for automatically generating HTML.}
\centering
\begin{tabular}{llp{8cm}}
\toprule
Prompt & \multicolumn{2}{p{10cm}}{You are a web developer. Please generate the HTML code for a web page with a caption of \{Deign\_Idea\}.} \\
\midrule
Slot  & Deign\_Idea  & Human-authored descriptions of web design ideas, with an example shown in \cref{tab:design_idea}.   \\
\bottomrule
\end{tabular}
\label{tab:template_HTML}
\end{table}

\begin{figure*}[h]
\centering
\includegraphics[width=0.99\textwidth]{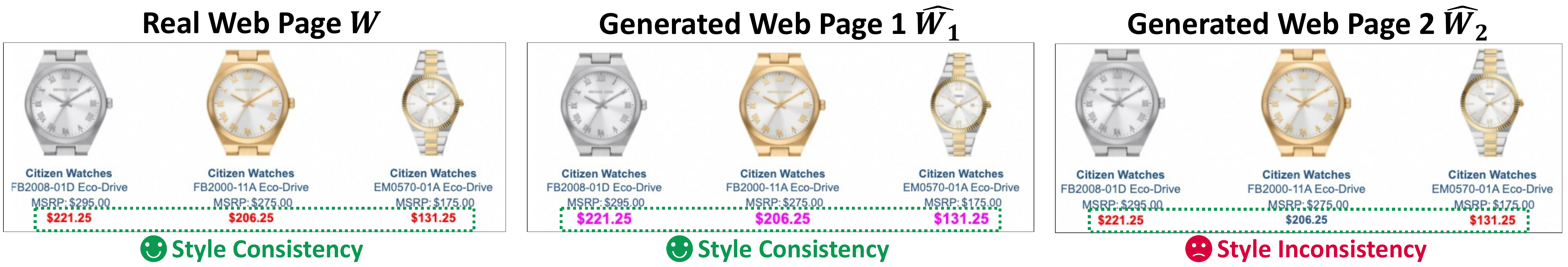} 
\caption{An example for visualizing style consistency. Notably, $\hat{W_1}$ and $\hat{W_2}$ are artificially created for demonstration purposes.}
\label{fig:SC_score_explain}
\end{figure*}

\begin{figure*}[h]
\centering
\includegraphics[width=0.8\textwidth]{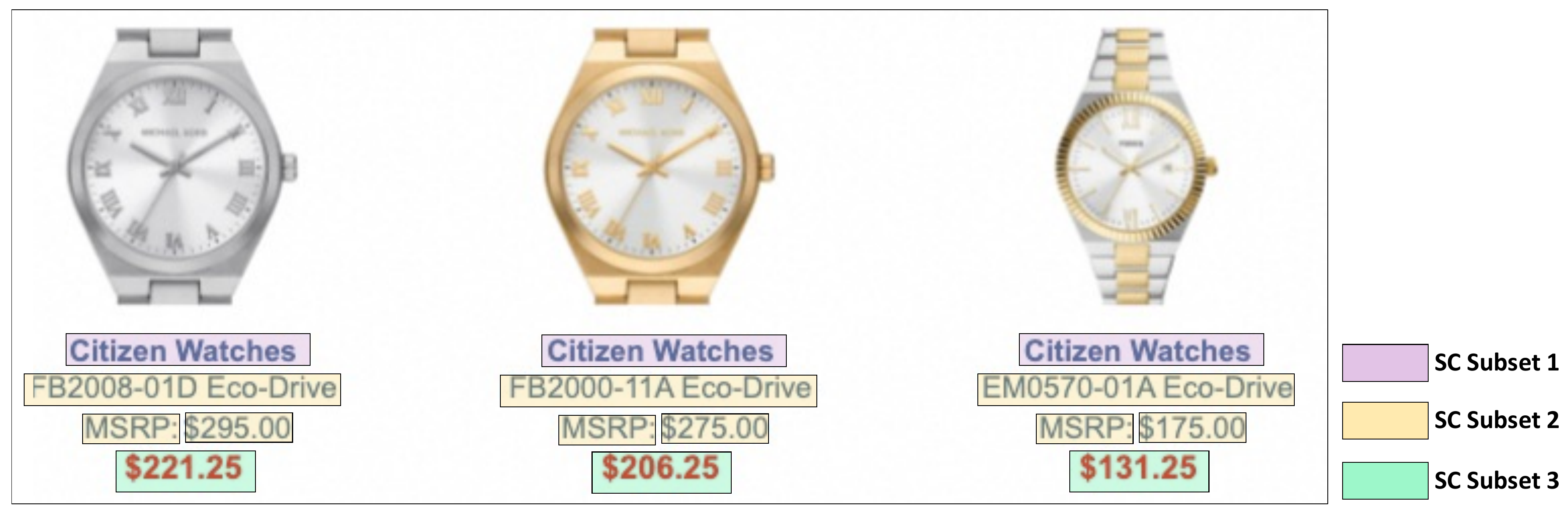} 
\caption{A visualization of the style consistency subset based on a real web page. The style consistency subset is defined in \cref{sec:sc_score}.}
\label{fig:SC_subset}
\end{figure*}

\subsection{Human Evaluation}
\label{sec:human_eva}
We conduct a human evaluation using pairwise comparisons. We randomly select 100 test samples and generate visual presentations using WebRPG-AR, WebRPG-DM, and GPT-4. Five human annotators evaluate each pair to determine the superior presentation or if there is a tie. The results, shown in \cref{fig:human_eval}, align with the objective evaluations in \cref{tab:results}.

\begin{table}[h]
\centering
\caption{An example of web design ideas described by humans.}
\begin{tabular}{p{9.5cm}}
\toprule
This web page showcases the ``Rumble Band for 38mm Apple Watch,'' offered at \$19.99. It's identified as the X-Doria Rumble Band and is noted for its compatibility with the 38mm Apple Watch Series 1, 2, 3, and Nike Edition. Highlighted on the page are customer assurances including a lifetime warranty, complimentary shipping on all orders, and a 30-day hassle-free return policy. A conspicuous ``Add to Cart'' button is prominently displayed. The product's image is designed to highlight its appearance and design features. \\
\bottomrule
\end{tabular}
\label{tab:design_idea}
\end{table}

\section{An Example Explanation of SC Score}
\label{sec:example_sc_score}

\cref{fig:SC_score_explain} provides an example to explain the SC Score further. The elements representing price (marked with a green box, hereafter termed as price elements) on the real web page \(W\) and on generated web page 1 \(\hat{W_1}\) have differing styles in terms of font color and size. However, these differences do not affect the perception of price elements, as their style remains consistent within each individual web page. In contrast, the generated web page 2 \(\hat{W_2}\) changes just one price element, which leads to confusion when perceiving the price elements. Although \(\hat{W_2}\) seems more visually similar to \(W\) because of only one differing element, from a semantic perspective, \(\hat{W_1}\) is more coherent. Therefore, the SC Score evaluates whether elements that share a style on the real web page maintain that consistency on the generated page, beyond just visual similarity. Additionally, \cref{fig:SC_subset} provides a visualization of the style consistency subset for a real web page.

\section{Further Discussion on the Performance of LLM in WebRPG Task}
\label{sec:LLM_in_WebRPG}

\begin{figure*}[h]
\centering
\includegraphics[width=0.99\textwidth]{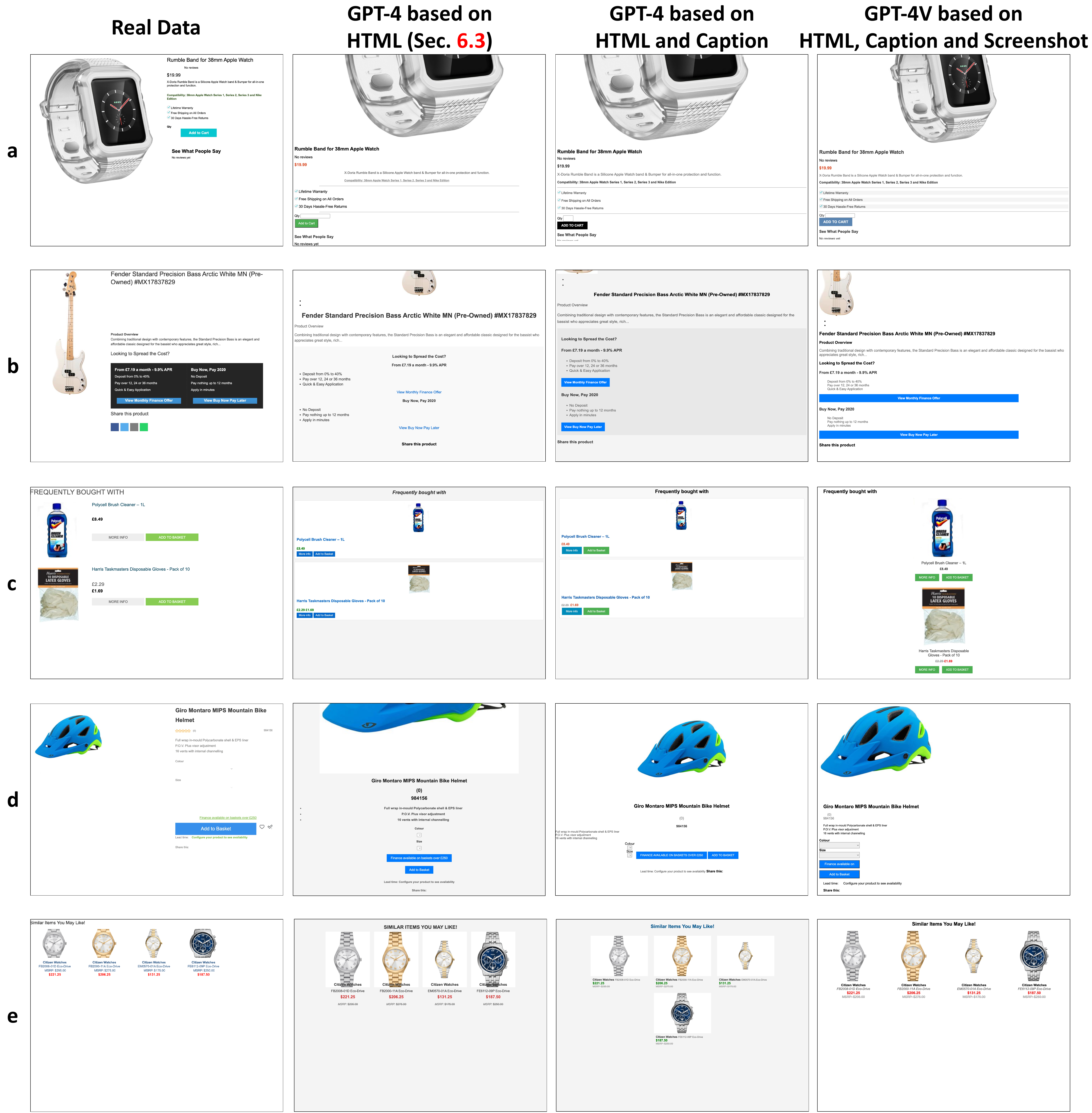} 
\caption{Further qualitative evaluation of GPT-4's performance in WebRPG task. Notably, the ``GPT-4 based on HTML'' group is the experiment in \cref{sec:experiment_result}.}
\label{fig:GPT_Gen_image}
\end{figure*}

As described in \cref{sec:implementation}, we employ GPT-4 as a representative for LLMs. Due to the complexity of CSS code practices and the noise in actual web pages, directly fine-tuning LLMs is not feasible. Consequently, we do not conduct fine-tuning experiments. Moreover, to further explore the performance of GPT-4 in WebRPG tasks, we conduct two qualitative experiments. \cref{tab:template} details the prompt templates. The first experiment inputs HTML and the captions from the original web page screenshots. The second experiment comprises HTML, these captions, and the screenshots themselves. It's noteworthy that the additional data comprised visual information from the original web pages, serving essentially as a form of ground truth. The second experiment and the generation of web page screenshot captions both leverage the multimodal capabilities of GPT-4V\footnote{\url{https://openai.com/research/gpt-4v-system-card}}. \cref{fig:GPT_Gen_image} presents visualizations of selected cases, showing that additional data does not enhance GPT-4's performance. Given that these two qualitative experiments involve ground truth inputs, we do not include them in the main text or conduct quantitative experiments.

\begin{table*}[h]
\centering
\setlength\tabcolsep{5pt}
\caption{The prompts for \cref{sec:LLM_in_WebRPG}. ``H.'', ``C.'', and ``S.''  denote ``HTML'', ``caption'' and ``screenshot'', respectively.}
\begin{tabular}{llp{7.5cm}}
\toprule
Information &
  H.+C. &
  You are an exceptional web designer. Please create the corresponding CSS code based on the HTML code I have provided, so  as to craft a well-designed visual presentation for the web page.   \textbf{Furthermore, for better comprehension of the original web page design, here is a detailed caption: \{Caption\}.} You can only use the following CSS   properties: "left", "top", "width",   "height", "font-style", "font-weight",   "font-size", "line-height", "color",   "text-align", "text-decoration",   "text-transform", "background-color". Please exercise caution in controlling the size of the image, as using the original image dimensions directly may result in excessive spatial occupation.   Here are several demonstrations:\{Demonstrates\}. Below is the HTML code and do not reply with anything other than CSS code: \{HTML\_Code\}. \\
 &
  H.+C.+S. &
  You are an exceptional web designer. Please create the corresponding CSS code based on the HTML code and \textbf{screenshot} I have provided,   so as to craft a well-designed visual presentation for the web page.   \textbf{Furthermore, for better comprehension of the original web page design, here is a detailed caption: \{Caption\}. }You can only use the following CSS   properties: "left", "top", "width",   "height", "font-style", "font-weight",   "font-size", "line-height", "color",   "text-align", "text-decoration",   "text-transform", "background-color". Please exercise caution in controlling the size of the image, as using the original image dimensions directly may result in excessive spatial occupation. Here are several demonstrations:\{Demonstrates\}. Below is the HTML code and do not reply with anything other than CSS code: \{HTML\_Code\}. \\
\midrule
Slots & Caption      & Captions from the original web page screenshots.          \\
      & HTML\_Code   & HTML code of given web page.                              \\
      & Demonstrates & The HTML-CSS pairs for three selected web page segments. \\
\bottomrule
\end{tabular}
\label{tab:template}
\end{table*}